\documentclass[final, 5p, authoryear, twocolumn]{elsarticle}

\usepackage{hyperref}
\usepackage{natbib}
\usepackage{amsmath}
\usepackage{amsfonts}
\usepackage{listings}

\journal{Arxiv}

\biboptions{authoryear}

\begin{document}

\begin{frontmatter}

\title{A new interpretable unsupervised anomaly detection method based on residual explanation \tnoteref{mytitlenote}}

\tnotetext[mytitlenote]{This work was supported by the Vale S.A, Vitória, Brazil [grant number 4600043577]}

\author{David F. N. Oliveira\fnref{address1}\corref{mycorrespondingauthor} }
\cortext[mycorrespondingauthor]{Corresponding author}
\ead{davidoliveira@usp.br}

\author{Lucio F. Vismari  \fnref{address1}}
\ead{lucio.vismari@usp.br}

\author{Alexandre .M. Nascimento  \fnref{address1,address2}}
\ead{alexandremoreiranascimento@alum.mit.edu}

\author{Jorge R. de Almeida Jr \fnref{address1}}
\author{Paulo S. Cugnasca  \fnref{address1}}
\author{João B. Camargo Jr \fnref{address1}}
\author{Leandro Almeida \fnref{address3}}
\author{Rafael Gripp \fnref{address3}}
\author{Marcelo Neves \fnref{address3}}

\address[address1]{School of Engineering, University of São Paulo (USP), São Paulo, SP, Brazil}
\address[address2]{Stanford University, Palo Alto, USA}
\address[address3]{VALE S.A, Vitória, Brazil}

\begin{abstract}
Despite the superior performance in modeling complex patterns to address challenging problems, the black-box nature of Deep Learning (DL) methods impose limitations to their application in real-world critical domains. The lack of a smooth manner for enabling human reasoning about the black-box decisions hinder any preventive action to unexpected events, in which may lead to catastrophic consequences. To tackle the unclearness from black-box models, interpretability became a fundamental requirement in DL-based systems, leveraging trust and knowledge by providing ways to understand the model’s behavior. Although a current hot topic, further advances are still needed to overcome the existing limitations of the current interpretability methods in unsupervised DL-based models for Anomaly Detection (AD). Autoencoders (AE) are the core of unsupervised DL-based for AD applications, achieving best-in-class performance. However, due to their hybrid aspect to obtain the results (by requiring additional calculations out of network), only agnostic interpretable methods can be applied to AE-based AD. These agnostic methods are computationally expensive to process a large number of parameters. In this paper we present the RXP (Residual eXPlainer), a new interpretability method to deal with the limitations for AE-based AD in large-scale systems. It stands out for its implementation simplicity, low computational cost and deterministic behavior, in which explanations are obtained through the deviation analysis of reconstructed input features. In an experiment using data from a real heavy-haul railway line, the proposed method achieved superior performance compared to SHAP, demonstrating its potential to support decision making in large scale critical systems.
\end{abstract}


\end{frontmatter}

\section{Introduction}
In recent years, advances in deep learning (DL) have enabled themselves to be widely adopted in real-world applications. As a prime representative, the Deep Neural Networks (DNN) have been achieving impressive performance  and reaching the state-of-the-art performance in diverse application domains such as computer vision, speech recognition and natural language processing ~\citep{lecun2015deep}. These achievements drove Machine Learning (ML), in particular the DL, to become a growing research topic due to its potential in automating activities dependent on higher levels of cognition previously performed only by humans. 

The application of DL techniques leads to structurally complex models. These complex structures preclude intuitions about the rationale behind the predictions, undermining the model trust. Consequently, DL-based models are adopted as black-box approaches by their end-users. However, this lack of transparency on how the results are obtained stand as a challenge for the adoption of these models in some real-world applications. Depending on the domain, some aspects are critical, such as safety (trust by robust reasoning in risky situations); fairness (equal opportunities ensuring no gender bias or racism) ~\citep{doshi2017towards}; and accountability (capable to justify the decision making). Besides, critical systems demand their decision-making models to learn and be able to apply reasonable rules in order to avoid unacceptable behavior that may lead to ethical/legal issues, profit losses or accidents.

Therefore, understanding how a decision is made by a ML-based model is fundamental in critical domains. For these reasons, Interpretable Machine Learning has received growing attention by the scientific community, given that it deals with the intrinsic transparency limitations in black-box models. Interpretability can be defined as the ability to enable intuition and reasonability about model output ~\citep{chakraborty2017interpretability} as a means to augment trust ~\citep{lipton2018mythos}. Consequently, interpretable models are a class of methods that use ML-based models to extract relevant knowledge from the data ~\citep{murdoch2019interpretable}, enabling clearness and insights about results obtained from black-box models. 
In general, interpretability methods are based on either the perturbation of the model’s inputs (perturbation-based) or the analysis of the model’s neuron activation/gradient behavior (gradient-based, exclusive for neural networks). Perturbed-based methods evaluate the impact on the output regarding different combinations of deviated samples in a particular input. The common manner to obtain explanations is by building \textit{proxy models} ~\citep{gilpin2018explaining}, that is,  a shallow model (such as linear regression or decision tree) that locally approximate the black-box behavior in the region of the particular input to be predicted. This shallow model is easily interpretable due its simple structure. Thus, perturbation-based methods are agnostic with respect to the predictive models by only requiring access to the model’s boundaries (input-output tuple), enabling them to be applied to any ML-based model. 

However, the trade-off between interpretation performance and response time is a inherent challenge due to the combinatorial aspects in order to find optimal results. In cases where the model input has a large number of parameters, one single record may require a large processing time to achieve more accurate interpretation results. In the same way, faster results can be obtained by sacrificing interpretation performance due to the randomness brought by searching for results in a smaller dimensional space. Consequently, perturbed-based methods are prohibitive in many real-time, large-scale applications. Examples of well-known perturbed-based methods are LIME (Local Interpretable Model-agnostic Explanations) ~\citep{ribeiro2016should} and SHAP (SHapley Additive explanation) ~\citep{NIPS2017_7062}. 

On the other hand, the gradient-based methods analyze neural network behaviors by looking for which network nodes are more excited (or deviated from the normal behavior), given a particular output. Gradient-based approaches are not negatively impacted by the size of inputs since they only evaluate a specific observation, resulting in a fast response. Furthermore, these methods deliver the exact same interpretation result for a given network and to the same input/output (deterministic behavior). Consequently, they are more suitable for real-time, large-scale purposes, allowing them to obtain robust interpretation results during runtime. Examples of this technique are DeepLIFT (Deep Learning Important FeaTures)~\citep{shrikumar2017learning} and Integrated Gradients~\citep{sundararajan2016gradients}. 

Despite gradient-based methods be more suitable for real-time, large-scale models, they are not agnostic. Thus, they are restricted to neural networks (i.e. DL-based models), in which the input-output relationship is raised by checking detailed behaviors of neuronal activation during predictions. In all other ML-based applications, the perturbed-based methods are the only option for interpretability, which are computationally inefficient to deal with a large number of parameters. A very typical ML-based application is the Anomaly Detection (AD) in large scales systems. AD is used to identify items or events that deviate from an expected behavior ~\citep{goldstein2016comparative}. AD is a common challenge in diverse domains, such as finance (fraud detection), network security (intrusion detection), industrial (fault detection) and healthcare (patient health checking). Independent of the application domain, ML-based approaches need to be trained to learn what is an anomaly, in which supervised and unsupervised methods are common approaches adopted to this task. 

Regardless the superior performance of supervised methods training ML-based models, AD imposes restrictive training requirements concerning to the task specificity's, such as (i.) supervised approaches are prone to be biased to mask anomalies, due to anomalies are rare events ~\citep{liu2008isolation} and, consequently, the datasets are very imbalanced; and (ii.) anomalies may happen in a novel and unseen ways ~\citep{chalapathy2019deep}, meaning that the assumption that all possible anomalies are known and mapped is rarely true. Therefore, supervised approaches may be inadequate and fail prone when there is limited knowledge (i.e. insufficient number of examples or lack of all possible anomaly scenarios for training). On the other hand, by exploring deviations in data, unsupervised AD approaches are flexible and more efficient to detect anomalies, including the cases when the observations come in a novel form.  

Autoencoder (AE), a type of neural network architecture, is the core of current unsupervised DL-based models for AD applications ~\citep{chalapathy2019deep}, in which a substantial number of researches can be found for AD issues. Unsupervised AD applications using AEs are composed by two basic components: (i.) The AE (neural network), generating a deviation score (a.k.a. residuals) to the data input; (ii.) A binary classifier, which decides about the input value (true or false) based on AE residuals rates. Thus, given the AD’s Input-Output relationship is not completely implemented by a NN, the results obtained by unsupervised AD applications using AEs cannot be explained (interpreted) by gradient-based methods. Consequently, the only option is using agnostic interpretable models, such as LIME or SHAP, which faces randomness and time processing challenges. 

Therefore, this paper proposes a method to implement interpretability capabilities to AE-based unsupervised Anomaly Detection (AD) applications. This method stands out for its implementation simplicity, low computational cost, and deterministic behavior. Its explanations are obtained by means of the deviation analysis of reconstructed input features. Consequently, it is more suitable for AE-based AD applications in large-scale systems. 

This paper is organized as follows: section 2 gives a background information for the present research; section 3 presents our proposed method; section 4 presents an experiment to validate the proposed method in a real-world critical application; section 5 presents the concluding remarks of this work, as well as its possible future steps.

\section{Background}
This section provides background knowledge about keys concepts supporting this work: section 2.1 gives a brief overview about reconstruction-based methods and traditional design of AE for AD tasks; section 2.2 describe a intuition about SHAP (kernel explainer - the baseline method compared with RXP) and LIME (interpretability method extended by SHAP); section 2.3 highlights some related works that propose different approaches to deliver interpretations about their results.  

\subsection{Autoencoder for unsupervised AD (AE-based AD)}
Different manners are found in the literature to deal with AD problems, such as distance-based ~\citep{goldstein2016comparative}, density-based ~\citep{10.1145/3416921.3416934}, or custom heuristics ~\citep{liu2008isolation}. Among them, reconstruction-based approaches have gain attention in recent researches using DL architectures ~\citep{chalapathy2019deep}.

Reconstruction-based approach attempts to transform the data into different dimensions (usually smaller), and then inversely returning to the original format. As anomaly detection problems usually deal with unbalanced datasets (since anomalies are "rare events") ~\citep{liu2008isolation}, the philosophy of using reconstruction-based models understands that models are trained to reproduce normal observations with high performance. On the other hand, these models tend to fail to faithfully reproduce anomalies. Thus, these models discriminate samples by calculating a residual error between normal (usually lower values) and failure (higher values). Common examples of back-end methods for reconstruction-based models are Autoencoders (AE), Restricted Boltzmann Machines (RBM), and Principal Component Analysis (PCA). 

As a prominent example of DL architecture applied for AD problems, Autoencoder (AE) is a type of artificial Neural Network (NN) that learns to reconstruct the network input to its output using unsupervised training. This is typically done by compressing the input through lower dimension layers (encoder), and then decompressing (reconstructing) the original input to its original size (decoder). This architecture allows AEs to be used for efficient Dimensionality Reduction (DR) and generative models. AEs learn intrinsic aspects from the data, creating a reduced dimensional latent space that discards noises and unrepresentative data. 

\begin{figure}
  \includegraphics[width=\linewidth,height=130pt]{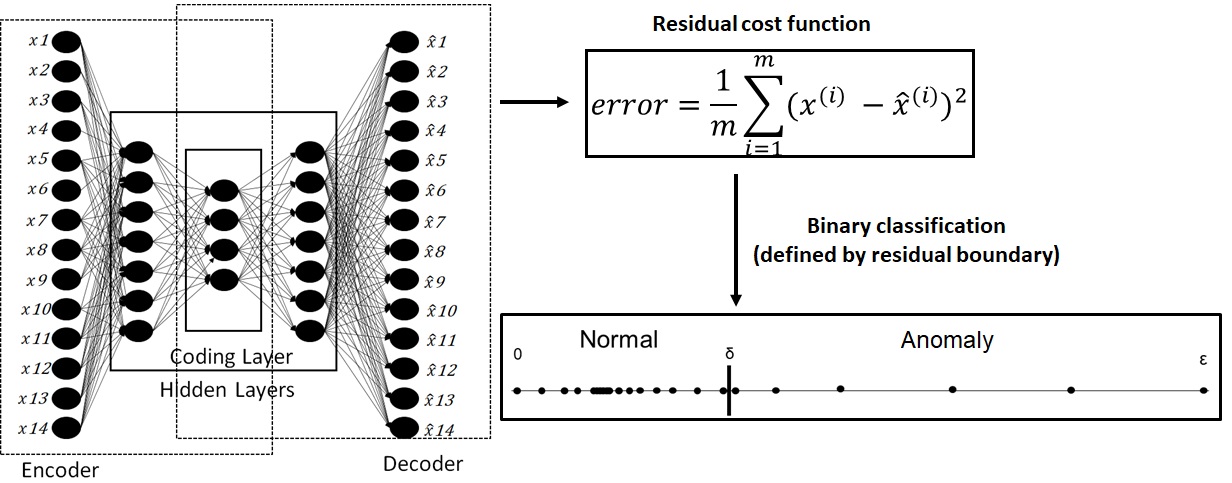}
  \centering
  \caption{Traditional approach for standard autoencoder applied to anomaly detection. Source: Authors}
  \label{fig:autoencoder}
\end{figure}

Figure \ref{fig:autoencoder} illustrates a traditional AE implemented as a feed-forward Multi-Layer Perceptron, typically applied to anomaly detection (AD). A multidimensional input is processed by artificial neurons through a sequence of nonlinear functions over multiple layers. This combined with the dimensional reduction over the layers results into rich dense representations in its hidden layers. The compressed data is then used to rebuild the original information from the lower representation. Many variations of AE can be found in the literature, such as Sparse ~\cite{chen2017multisensor}, Denoising ~\cite{vincent2010stacked}, Variational  ~\cite{kingma2013auto}, and hybrid combinations such Recurrent~\cite{marchi2015non} and Convolutional ~\cite{koizumi2018unsupervised}. 

Since AE for AD follows the reconstruction-based approach, are classified by calculating residual error in the reconstruction of the encoded input by the decoder. Residual values are commonly obtained by a mean squared error function, wherein for each parameter \textit{i}, $x_i$ is the original input, ${\hat{x}}_i$ is the reconstructed input. A contamination hyperparameter (percentage of faulty records in the data) should be defined to find the residual boundary value between normal and fault for the binary decision. During the training phase, the lowest record within the highest residual contamination percentage defines the residual threshold value (${\delta}$) for the binary classification between normal and anomaly. 

\subsection{Agnostic Interpretability Methods}
Among the manners to provide interpretations from black-boxes, model-agnostic interpretability methods stand out for the independence about intrinsic aspects of the predictive model to produce responses. Interpretations are commonly generated by forcing noises in the input and checking respective impacts on the outputs. Higher deviations may indicate that the perturbed attribute is more relevant to the particular prediction. Below we detail LIME e SHAP (kernel explainer), two of the most popular methods in this line.
\subsubsection{LIME}
\textit{LIME} ~\cite{ribeiro2016should} grounds on the assumption that complex black-box models can be simulated by shallow models in the region of the instance to be explained (also known as local surrogate methods). A white-box model is trained using generated perturbed samples around the original input. The optimization of the white-box method aims at maximum approximation as to the response of the black-box model in the region of the observation to be explained, which can be expressed as follows: 
{\it
\begin{equation}
        e\left(x\right)= \underset{g\ \in\ G}{argmin} \; {L}\left(f,g,\pi_x\right)+\mathrm{\Omega}\left(x\right)
\end{equation}
}wherein the function \textit{e} represents the best explanation for the reduction of the residual error \textit{L} (usually the quadratic error), between the black-box model \textit{f} and the approximate white-box model \textit{g}, belonging to the space of possible explanations \textit{G}, with control of parameters by regularization $\Omega$, and trained with a sample volume of neighbors $\pi_x$ (the more neighbors, the greater the quality of results). The white box methods are generated with the objective of locally approximating the prediction results of the black box methods, and provide intuitive means of interpretation, native for their structure, such as linear regression, which follows the form
{\it
\begin{equation}
        g\left(x\right)=\ \beta_0\ +\sum_{i=1}^{M}{\beta_ix_i}
\end{equation}
}in which the input \textit{x} composed of \textit{M} parameters and their respective weights $\beta$ (in which \ $\beta_0$ \ is the bias) are added. The parameter importance is assessed by of the module its value associated with weight ($\beta_ix_i$). Other types of white-box method can be applied, such as decision trees, and decision rules (IF-THEN). In addition, the method is prepared to receive different types of entries such as tabular records, free text or images.
\subsubsection{SHAP (Kernel Explainer)}
\textit{SHAP} ~\citep{NIPS2017_7062} is an interpretability method inspired by Shapley Values ~\citep{shapley1953value}, a solution concept on cooperative game theory. This game comprises of several participants (input features) that act with a common goal to be achieved (prediction), are rewarded fairly in relation to his contribution to gambling (feature relevance), being represented by the function
\begin{equation}
    \varphi_i\left(v\right)=\sum_{z\ \subseteq\ N\{i\}} 
    {
    \frac{\left|S\right|!
    \left(N-\left|S\right|-1\right)!}{N!}
    \left(v\left(z\right)-v
    \left(z_i\right)
    \right)
    }
\end{equation}

in which calculates the average contribution (Shapley Value) of a player (i.e. input feature) \textit{i} in relation to the total number of players (i.e. full set of features) \textit{N}. This is done by capturing his marginal contribution (i.e. feature relevance) on the difference obtained between his participation or not in different coalition permutations (subgroups) of players \textit{z}, calculated by function \textit{v}. The contribution of an ordered coalition of players is represented by $v\left(z\right)$, and $v\left(z_i\right)$ represents the same coalition without the contribution of player \textit{i}. The contribution of each player varies depending on the set of players that compose the coalition. A unique distribution of the average contribution of all game participants is created, through checking each of the possible permutations. The Shapley value obtained corresponds to the marginal contribution of each parameter, obtained fairly in relation to the importance of all parameters.

Lundberg (\citeyear{NIPS2017_7062}) raises three important properties delivered by SHAP that are consistent concerning Shapley Values, suggesting that the features relevance weighting are done fairly with a strong theoretical background (being a disregarded matter by LIME): 
(i.) \textit{local accuracy}, which similar to LIME, this method has the objective of local approximation about the behavior of a black box method regarding a specific observation, generating an explanation by the linear additive combination (similar to Equation 2) in form
{\it
    \begin{equation}
        g\left(x\right)=\ \varphi_0\ +\sum_{i=1}^{M}{\varphi_ix_i}
    \end{equation}
}in which the Shapley Value ($\varphi_i$) represents the weight of parameter i on the prediction of a point observation.
(ii.) \textit{lack of importance}, meaning in a simplified view of the reduced dimensional input, the lack of importance indicates the absence of a parameter without affecting the expected response; (iii.) \textit{consistency}, denoting that if the white box model is updated so that the contribution of one parameter increases or remains the same, regardless of the other parameters, its value must not be reduced.

In cases where the input data have an extensive number of parameters leads to a high computational cost over coalition permutations. SHAP suggests an optimized way to calculate important parameters using the conditional average of the original Shapley Values model. The author suggests that the premise of independence and linearity (Equation 4) between the parameters simplifies the computational complexity. This is done by reducing the cases to be evaluated (i.e. the evaluation of coalitions of parameters are not evaluated in different orders, only the inclusion of new parameters). Then, it demonstrates that a white box function can be approximated in a more efficient way.

SHAP deals with the error function stated by LIME (Equation 1), but defining specific patterns and adjustments on each of the components to minimize the error, 
which is called the \textit{SHAP Kernel Explainer}. It suggests well-defined methods for optimization (error minimization standards and adjustments generally suggested by LIME), promoting an improvement and efficiency gain for the generation of samples. This paper, Kernel SHAP is used on the experiments. The authors also suggest solutions aimed at specific types of models, which are not discussed in this work.
\subsection{Related works on interpretability methods to AE-based AD}
Some relevant works dealing with interpretability methods using AE can be found in the literature. ~\cite{antwarg2019explaining} propose a combination with SHAP using the ranking of top residual features obtained by the AE to be interpreted individually against the explainer. SHAP values are recalculated using the set of interpretations. ~\cite{shankaranarayana2019alime} present a method called ALIME, which combines denoising autoencoders and LIME by optimizing the sampling task through the weighting training instances using the latent space representations. Authors show ALIME improves stability on the results with a faster response time. 

~\cite{akerman2019vizads} detail an interpretable AD method on image streaming for automatic dependent surveillance-broadcast in air traffic control systems based on LSTM-Autoencoder. The proposed method is capable to identify frames and regions of the image with high accuracy by checking regions with high residual errors.  ~\cite{donahue2019deep} defines a method based on Generative Adversarial Networks (GAN) called AnoGAN that is trained with normal data to detect anomalies in images of X-ray CT scans. An anomaly interpretation is done by comparing normally generated images with real images with similar distribution, in which anomalous regions are identified by strong dissimilarity forms.  

~\cite{nguyen2019gee} use Variational Autoencoders (VAE) for network intrusion detection. Interpretations are extracted by applying clustering methods over gradient behavior on the latent representations, in which anomalies present different patterns from the normal.  ~\cite{park2019fault} propose the use of AE and Long Short-Term Memory (LSTM) neural networks for fault detection and diagnosis. AE captures incipient anomalous behaviors and its compressed layer outputs are used as dimensionality reduction strategy for LSTM networks on fault diagnosis tasks.

Thus, diverse AE architectures to compose interpretable models can be found in literature. Some works just use AE as auxiliary tools to optimize other predictive methods ~\cite{park2019fault}. Others use agnostic methods such as LIME ~\cite{shankaranarayana2019alime} and SHAP ~\cite{antwarg2019explaining}. Some other works focused on specific architectures such as AnoGAN ~\cite{donahue2019deep} and LSTM-AE ~\cite{akerman2019vizads}, made for specific application domains (spatial and temporal data input types). However, we can observe that all these propositions are prone to dimensionality constraints and, consequently, they are computationally expensive when a large number of parameters is used to as input to AD application. This lack of efficiency can jeopardize their practical use in large-scale AD applications. 

Therefore, interpretability methods to be used in AE-based AD applications that deal with abundant data availability with low computational cost are demanded. Thus, we propose a new interpretability method that deals with common tabular data (as usual in real industrial scenarios), being capable of delivering deterministic and near real-time interpretations, which makes it more suitable for practical purposes.

\section{The Residual eXPlainer (RXP) Method}
Considering the training set \textit{\(T\in\mathbb{R}^{N\times M}\)} having \textit{N} instances and \textit{M} features, statistical data (standard deviation and mean) are gathered from the reconstructed residuals for each parameter. This statistical data is used to calculate the Z-score, defined by the Eq.(\ref{eq:zscore}), in which \textit{\(z_{nm}\)} is the deviation score for each parameter \textit{\(m\)} belonging to the sample \textit{\(n\)}, \textit{\(x_{nm}\)} is the original input, \textit{\(u_{m}\)} is the residual training mean, and \textit{$\sigma_{m}$} is the residual training standard deviation. 

{\it
\begin{equation}
        z_{nm}=\frac{x_{nm}\ -\ u_{m}}{\sigma_{m}}
        \label{eq:zscore}
\end{equation}
}During the test phase, given one particular instance \textit{\(n\)} classified as anomaly, the  relevance \textit{\(R_{nm}\)} of a feature \textit{\(m \in M\)} is defined by Eq.(\ref{eq:proposed_method}), in which the nominator is the  residual cost function (squared error between the original \textit{\(x_{nm}\)} and reconstructed \textit{\({x^{'}}_{nm}\)} input) weighted by log module of Z-score \textit{\(z_{nm}\)} between the original feature input and its mean residual training value. This weighting module is smoothed by log to reduce the strong effect caused by binary features (when existing together with continuous features) that may lead to very imbalanced relevance attributions. The denominator acts as normalization factor along all feature dimensions. Higher values for \textit{\(R_{nm}\)} means that the attribute \textit{\(m\)} has higher relevance for the decision of the instance \textit{\(n\)} as an anomaly.
{\it
\begin{equation}
        R_{nm}= \frac{\log(1+|z_{nm}|)\left(x_{nm} - {x^{'}}_{nm}\right)^2 }
        {  \sum_{i=1}^{M}{\log(1+|z_{ni}|)\left(x_{ni}- {x^{'}}_{ni} \right)^2 }}
\label{eq:proposed_method}
\end{equation}
}Note this method takes into account not only  the deviations on the reconstruction, but considers the deviated input as well (raised by Z-score). Therefore, even for cases in which the network learns good reconstruction of some anomaly, the Z-score part tends to penalize deviated inputs relative to the training data. Consequently, this helps to mitigate possible mistakes that could hinder the interpretability task due to high-quality reconstructed patterns learned by the AD model. In such cases, the observations commonly have slight deviations from the normal behavior, and the AD model returns very low and balanced residual rates, bringing confusion to deliver accurate relevance weighting.  

\section{Experimental Evaluation}
AD is a useful approach to detecting faults in industrial applications context, since ‘faults’ are anomalies – rare events that deviate a system from its expected behavior ~\cite{avizienis2004basic}. Likewise as it has been deeper explored for patient health diagnosis in medical applications ~\cite{elshawi2019interpretability}, we claim that interpretability models can also be useful in the industrial context. This is feasible by delivering further information about the cause of a fault event (fault diagnosis) by raising which parameters were relevant to a fault detection classiﬁcation. Therefore, our experiments aim to compare RXP against Kernel SHAP as the fault diagnosis solution in a real scenario on machinery operation in the railway domain. In this section we present the case study (dataset), experimental process, evaluation metrics and results.

\subsection{Case Study}
We used an exclusive dataset obtained from multiple sensors deployed along the Vitoria-Minas Railway track ('Estrada de Ferro Vitória a Minas' - EFVM, Brazil), operated by the Brazilian mining company VALE S.A. 

\newcommand{\specialcell}[2][c]{%
  \begin{tabular}[#1]{@{}c@{}}#2\end{tabular}}
\begin{table*}
\caption{Three examples of parameters from different wayside equipment}\label{tab1}
\begin{tabular}{|c|c|c|c|c|}
\hline
Parameter & \specialcell{Wayside \\Equipment}. & \specialcell{Component} & \specialcell{Side} & \specialcell{Axle}\\
\hline
\lstinline $HEAT_WHEEL_LEFT_AXLE3$ & HBW & Wheel & Left &  3\\
\lstinline $RS_R_AXLE2$ & ABD &	Bearing & Right & 2\\
\lstinline $DIR_IMPACT_MAX_AXLE1$ & WILD &	Axle & - & 1\\
\hline
\end{tabular}
\centering
\end{table*}

Three types of wayside equipment were used to get measures from wheels and bearings of rail cars: Hotbox and Wheel (HBW) captures thermal emission from wheels and axle boxes (bearings) through infrared sensors; Wheel Impact and Load Detector (WILD) measures the force between the rail and moving wheels through the analysis of vertical load in the line; and Acoustic Bearing detector (ABD) captures acoustic emissions and classify possible faults in wheels and bearings. Totally, 15 HBW, 1 WILD and 1 ABD are deployed along EFVM, and they were used to get data with about 257 parameters. 

The raw data, obtained by wayside equipment, contains information at axle level (4 axles per rail car). Table 1 shows an example of parameters captured by each wayside and its respective links to the components and regions on the railcar. Since many information from the same equipment (railcar) is spread out over different datasets (waysides) and records (per axles), all records were grouped to rail car level, encompassing all axles in an unique record per track region. This approach enriches the case study by allowing to evaluate the capacity of the method to explain faults in holistic view by checking different causes (overheating, vibration or acoustic), components (wheel, bearing or axle), and regions (left/right side, axle number) of the same rail car.

\subsection{Experimental Protocol}
\begin{figure}
  \includegraphics[scale=0.50]{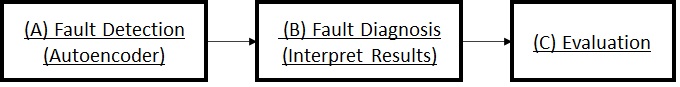}
  \centering
  \caption{The 3-step experimental process}
  \label{fig:methodology}
\end{figure}

This experiment was performed using a 3-step process, as illustrated in Fig. \ref{fig:methodology}. In \textbf{Step (A)}, the AE model were trained for the fault detection task. We followed the hyperparameter tuning defined by ~\citeauthor{oliveira2019evaluating}: large number of layers (20); high compression on latent layers (up to 98\%); no use of dropout/batch normalization; mean squared error for cost function and residual analysis; hyperbolic tangent as activation for latent layers and sigmoid for output. This model were used to detect faults (binary classification) and their respective residual rates for the diagnosis step (step B).

In \textbf{Step (B)}, the interpretability methods is used to obtain the explanations (by means of top-weighted relevant parameters per record) to the faults detected by step A. The proposed method (RXP), Eq.(\ref{eq:proposed_method}), and SHAP ~\cite{NIPS2017_7062} were applied and their results are compared in step C. Due to its deterministic behavior (giving the pair input/output), RXP is tested just once per sample set. On the other hand, SHAP may take longer to obtain consistent results when applied in problems with a large number of parameters. Therefore, 3 different scenarios are defined for SHAP, exploring the relationship between results precision and response time. The larger the size of the search space the less random the results, but higher is the response time. On the other hand, the smaller the size of the search space the more prone to unexpected or inconsistent results, but faster is the response time. Explanation results from this step were evaluated in step C.

\textbf{Step (C)} evaluated the results from each interpretability method. Step B returned a ranking of relevant parameters with their respective weights about the classiﬁcation. We checked if the expected parameters are ranked in top-weighted, and be capable of evaluating whether the diagnosis returns closer answers relative to the exact expected one. Each interpretability record was evaluated by computing a \textit{ranking-based Mean Average Precision (MAP) score} (Eq. (\ref{eq:average_precision})) over the top relevant parameters. This metric recursively raises the precision for each ranking position, starting from the top to the bottom. It captures the concentration of the expected results on the top and penalize any absent expected.

{\it
\begin{equation}
        MAP=\frac{1}{n}\sum_{n}\sum_{k}{{(S}_{nk}-\ S_{nk-1})\ P_{nk}}   
        \label{eq:average_precision}
\end{equation}
}$S_nk$ is the recall (sensitivity) and $P_nk$ is precision for the threshold at the top $K$ relevant features for query $n$. Recall is a metric used to evaluate how capable an AD method is to identify correctly (positively) the 'normal' records in a dataset (i.e. the true positive rate, or TP/(TP+FN)). TP is the number of True Positive results (correctly identified as normal by the AD) - in our case, be part of top-weighted K relevant parameters - and FN is the number of false negative results (relevant parameters wrongly classified as not important). Precision check the frequency of TPs relative to all samples classified as positives (TP/(TP+FP)), in which FP is the false positive cases (not relevant features wrongly classified as relevant).

\subsection{Results and Discussion}

For the fault detection step (A), the AE was trained using around 3.5 million samples in 1 epoch, 32 samples per batch, 25 layers with compression up to 98\% between the raw dimension (257) and the coding layer (5). Further details about the other hyperparameters can be found in ~\cite{oliveira2019evaluating}. The AE was tested against around 379 thousand records containing 3867 faulty samples, achieving a precision (P) of \textbf{82.5}\% and recall (S) of \textbf{95.3}\%.

\begin{table*}[t]
\caption{Experimental Results (SHAP1, SHAP2, SHAP3 and RXP).}\label{tab1}
\begin{tabular}{|c|c|c|c|c|}
\hline
Overall Results & SHAP1 & SHAP2 & SHAP3 & RXP\\ 
\hline
MAP (Eq. (\ref{eq:average_precision})) & 80.47\% & 80.61\% & 79.54\% & 81.38\% \\
\hline
Mean response time (ms) & 11,100 & 5,650 & 170 & 0.272 \\ 
\hline
Paired T-test & $3.3\times10^{-10}$ & $6.5\times10^{-10}$ & $3.2\times10^{-12}$  & - \\ 
\hline
\end{tabular}
\centering
\end{table*}

\begin{figure}
  \includegraphics[width=\linewidth,height=230pt]{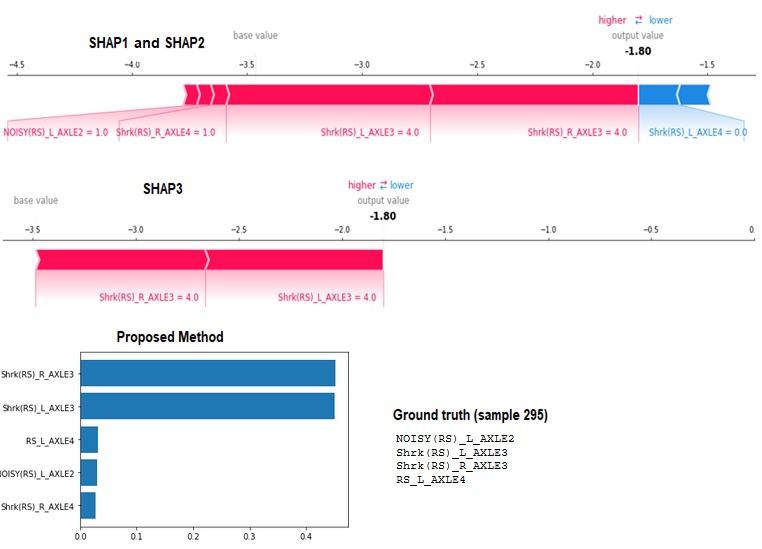}
  \centering
  \caption{Interpretation results of sample '295' containing 4 fault causes. Source: Authors}
  \label{fig:interpretation_example}
\end{figure}

The dataset composed by all the faults detected by step A (TP + FP) and false negatives cases (FN) were considered by step B. We performed 30 tests with 200 different random samples (with replacement) in order to reduce any random behaviors obtained from SHAP scenarios. These samples were interpreted by our proposed method (RXP) and SHAP in 3 different scenarios (SHAP1, SHAP2 and SHAP3). SHAP1 was tuned to return more precise results with low responses (around 7 seconds per record) by generating \textbf{800} simulated samples using \textbf{200} background training examples. SHAP2 was tuned for faster responses (around 5 seconds per record) by generating \textbf{800} simulated samples using \textbf{100} background training examples. SHAP3 was tuned to deliver very fast results (around 150 milliseconds per record) by generating \textbf{80} simulated samples using \textbf{10} background training examples, but presenting very unstable interpretation results. Figure \ref{fig:interpretation_example} shows an example of a  true positive sample containing 4 fault causes related to ABD wayside. SHAP1, SHAP2 and RXP were able to identify all the relevant parameters with strong weight importance (comprised by the larger bars on each plot), while SHAP3 fail to identify 2 fault causes. 

Results obtained by this experimental process - Mean Average Precision (MAP), Mean Response Time (in miliseconds), and the paired T-test from results (RXP versus SHAP) - are presented in Table 2. Experimental results show that RXP was faster than the SHAP scenarios, achieving up to 625 times faster relative to SHAP3, 20,772 times faster than SHAP2, and 40,845 times SHAP1. Besides, our method achieved the best results regarding the mean average precision (MAP).The RXP delivers superior performance relative to SHAP models, achieving p-value less than 0.00000001\% on paired T-test.

\section{Conclusions}
In this work, we propose a novel interpretability method for AE-based AD tasks through the reconstruction residual analysis called RXP. We compared our method with SHAP using real data from railway operations. The experiment compared the methods by simulating a fault diagnosis task, raising the relevant parameters that lead a sample to be classified as faulty (anomaly). For practical purposes, SHAP presented good performance over MAP metric but suffers from the trade-off between delivering consistent results (mitigate random behavior) and cost of evaluating large combinatorial parameter sets. On the other hand, RXP showed superior results, delivering near-real time response with high precision. Moreover, we focused to check the behavior of RXP over different sample tests relative to SHAP. As a key factor in critical  systems, other experiments must be  performed by running several tests over the same sample set to highlight possible instabilities caused by the non-deterministic nature of SHAP.

Through the weighted Z-score, our method also becomes sensitive to deviations even on instances with no gross deviation over any particular parameter (but slightly perturbed in a set of them). In this work, we considered the global residual calculation (squared error overall dimensions) and their respective weight of each parameter on the value. Further experiments are demanded in order to check appropriate usage for global residual calculations or relative to the parameter in other scenarios.   


\section*{Acknowledgments}
The authors would like to thank VALE S.A. for the institutional support that made this work possible.

\bibliography{ref.bib}

\end{document}